# Next word prediction based on the N-gram model for Kurdish Sorani and Kurmanji

* Hozan K. Hamarashid[1], Soran A. Saeed[2] and Tarik A. Rashid[3]

[1]*Computer Science Institute, Sulaimani Polytechnic University, Sulaimani, Iraq*

[2]*Vise President, Sulaimani Polytechnic University, Sulaimani, Iraq*

[3]*Computer Science and Engineering Department, School of Science and Engineering, University of Kurdistan Hewler, Erbil, Iraq*

[1]**Hozan.khalid@spu.edu.iq**, [2]*Soran.Saeed@spu.edu.iq*, [3]*Tarik.Ahmed@ukh.edu.krd*

*Corresponding Author is: Hozan K. Hamarashid*

**Abstract**

Next word prediction is an input technology that simplifies the process of typing by suggesting the next word to a user to select, as typing in a conversation consumes time. A few previous studies have focused on the Kurdish language, including the use of next word prediction. However, the lack of a Kurdish text corpus presents a challenge. Moreover, the lack of a sufficient number of N-grams for the Kurdish language, for instance, five grams, is the reason for the rare use of next Kurdish word prediction. Furthermore, the improper display of several Kurdish letters in the Rstudio software is another problem. This paper provides a Kurdish corpus, creates five, and presents a unique research work on next word prediction for Kurdish Sorani and Kurmanji. The N-gram model has been used for next word prediction to reduce the amount of time while typing in the Kurdish language. In addition, little work has been conducted on next Kurdish word prediction; thus, the N-gram model is utilized to suggest text accurately. To do so, R programming and RStudio are used to build the application. The model is 96.3% accurate.

## 1. Introduction

The Kurdish language is among a group of Indo-Iranian languages. Kurdish people speak in the Kurdish language. Kurdistan is geographically known as a mountainous area that consists of southeastern Turkey, northwestern Iran, northern Iraq, and northern Syria. The Kurdish Sorani dialect is written from right to left, in contrast to the English language. The Kurdish language consists of thirty-four letters and includes various dialects, for instance, Sorani and Kurmanji. Iraqi and Iranian Kurdish people speak in Kurdish Sorani (MacKenzie, 1961). The Kurdish language is constructed of vowels and consonants. Vowels can be short or long, and Kurdish language has ten vowel phonemes. Consonants can take different forms depending on where they appear in the word. The writing structure for Kurdish Sorani is from right to left, similar to Arabic script, but Kurmanji uses Latin script, which goes from left to right. The grammar rules of the Kurdish language include the word order, nouns, absolute, indefinite, and definite states. Generally, word arrangement or orders contain the subject, object, and verb.

Although the writing in the Kurmanji dialect is similar to that of the English language, the writing of the Sorani dialect is completely different from that of English in terms of characters and the morphology of the letters. In addition, the Sorani dialect is written from right to left. More details on the Kurdish language are given in section 3.

The next word prediction system is utilized in different fields, including mobile applications and word correction and word spell check systems (Nagalavi and Hanumanthappa, 2016). Accordingly, next word prediction is crucial and has many advantages; it is useful for people with dyslexia. Additionally, people who consistently misspell words could benefit from this technology. Furthermore, the prevalence of next word suggestion suggests that it can be used on various platforms.

The proposed application is constructed for the Kurdish language (Sorani and Kurmanji dialects). In this system, when a user writes a word, the next five words will be suggested to the user. That is, the proposed system will suggest the next five words based on the preceded written word or words. These suggestions depend on a provided Kurdish text corpus and use the N-gram model. The N-gram is mainly a sequence of *N* items or words (Kumar, 2017) and depends on the probability of word conflation occurrence based on a provided text or corpus.

The N-gram technique is utilized for the proposed system. Initially, a Kurdish text corpus was provided. Then, this corpus undergoes data preprocessing. In other words, the corpus is processed to be cleaned; for instance, non-Kurdish words are omitted, all the numbers are removed, symbols such as (", ', /), etc. are eliminated, and all punctuation marks and unnecessary white spaces are eliminated. Consequently, all Kurdish words remain in the corpus. After that, the corpus is prepared in order to make N-grams. To do so, the R programming language and R Studio are used. The corpus is read via the software. Later, it is converted to UTF-8 such that the words cannot be read properly until they have been encoded to UTF-8. However, by saving N-grams as an object, several Kurdish characters, such as (پ، ژ، ڵ، ێ، ۆ، ە), cannot be shown; they are shown as (U+0647, U+0695) and so on. They have to be replaced with their Kurdish letters, which are addressed later. The corpus is tokenized to make N-grams, for example, (a unigram, bigram, and trigram), and so forth. The constructed N-grams were then saved as an object; nonetheless, as mentioned above, several letters are still unreadable, as they are shown as codes (i.e., U+0647). As a consequence, all the saved objects are exported to Excel files. Next, in the Excel files, all the encoded characters are replaced with their Kurdish letters. Then, all the Excel files are read again by the software. Each Excel file contains an N-gram file where *n=1, 2, 3,* etcetera. Each Excel file consists of words or words and their frequencies. Last, the N-gram files are input through the N-gram model, and based on the N-gram model, the predictive words are revealed.

The contributions of this work are the following. First, it builds the first Kurdish corpus text, which includes approximately 500,000 words for both dialects. Second, it solves the problem of reading Kurdish characters in Rstudio, as the software does not read Kurdish characters (پ، ژ، ڵ، ێ، ۆ، ە). These letters have been encoded to UTF8, which produces encoded characters (i.e., U+0647). After that, the encoded character is replaced with the related character in the Kurdish letter (ە), etc. The main contribution is that next word suggestion has been developed for the Sorani and Kurmanji dialects of the Kurdish language for the first time.

The rest of this paper is organized as follows. A literature review is presented in section 2. In section 3, the Kurdish language is described. The algorithm and theory of N-gram are discussed in detail in section 4. The proposed system is explained in section 5. The results and simulations are established and analysed in section 6. Finally, the conclusion is outlined in section 7.

## 2. Literature Review

To date, very few studies have focused on the Kurdish language. Text classification for Kurdish documents has been achieved by utilizing different techniques and assessing the results. Researchers have developed accurate supervised text categorization, but for the Kurdish language, text categorization is extremely difficult; challenges include complicated morphologies of the Kurdish language. The reason beyond this difficulty is the enormous utilization of inflectional and derivational affixes (Rashid et al., 2018; Saeed et al., 2018; Mustafa & Rashid, 2017; Rashid et al., 2016). In addition, there are challenges associated with writing in the Kurdish Sorani dialect, which commonly uses suffixes and possessive pronouns (Rashid et al., 2018; Saeed et al., 2018; Mustafa & Rashid, 2017; Rashid et al., 2016). Building lemmatizers and spell checkers for Kurdish Sorani is possible because of the expansion of digital text. The stemming system can ne applied to the Kurdish language (Salavati et al, 2013). The main objective of stemming and lemmatization is to diminish various structures of words, namely, to determine the roots of words. In the stemming system, obtaining the base form of the word, which is called the stem, is procured by excerpting and trimming the derivationally associated form of the word. In different circumstances, lemmatization focuses on the articulation of a given word to scale it down to a dictionary form. The spell checker algorithm is one of the most significant applications of lemmatizers (Salavati et al, 2013). If an incorrect word is potentially given in a spell checker system, then the proposed system will provide a list of valid corrections. The Renus system for Kurdish Sorani proposes a spell checker. It uses the N-gram model with the rule-based method despite Kurdish Sorani's morphological rules (Shahin Salavati, Sina Ahmadi, 2018). This model has been utilized in a variety of aspects, especially in grammatical correction (Hernandz and Calvo, 2014).

Stemming for information retrieval in the Kurdish language has also been implemented. A rules-based stemmer has been used in an application. Rules-based stemmers implement a group of alteration rules to each word and attempt to band its suffix (Salavati et al, 2015). In contrast, a statistical stemmer has been utilized and addresses languages independently, which means that it demands a corpus only. Based on the aforementioned approaches, Kurdish stemming was manipulated by using the Kurdish stemming rules-based and statistical approaches. Consequently, various papers have been written in the Kurdish language. All the abovementioned works, however, differ from our work because none of them relates to next word prediction. In the following section, the Kurdish language and its alphabet are addressed.

## 3. Kurdish Language

The Kurdish language is similar to any other language, as it has its own letters or characters. However, the script in the Kurdish language is similar to Arabic, Persian, and Urdu. It consists of different dialects, such as Sorani (Arabic script-based) and Kurmanji (Latin script-based). Kurdish people in Iraq and Iran (MacKenzie, 1961) speak Sorani. The Kurdish alphabet for both dialects is shown in Figure 1:

Figure 1: Esmaili et al, 2013, Sorani Kurdish versus Kurmanji Kurdish

The writing system in Kurdish Sorani is from right to left, and vowels are nearly written as isolated letters; this contrasts with the original Arabic script, while other writing systems are enhanced by it. Namely, diacritics are used to represent certain or short vowels by positioning them above or under the letters and sometimes are deleted or excluded. Another point is that writing in Sorani is different from the Arabic script-based system. Arabic letters express some sounds that do not exist in Kurdish Sorani; sometimes, they are replaced by other letters to better express Kurdish pronunciations. For instance, the Arabic word طاقه (/tɑqa) is normally written in Sorani as تاقه, which involves replacing the letter ط (/t/?) with the letter ت (/t/). There are four other letters in Sorani that are used, such as ژ (/ʒ/), چ (/t͡ʃ/), گ (/g/), and پ (/p/). They do not exist in original Arabic characters, but they are utilized in the Persian alphabet. Additionally, the system for writing in Sorani does not include Tasheed, which is included in the Arabic writing system. On the other hand, in some cases, double consonants are discovered, in which a consonant is written twice, for example, وەڵڵا (/Wɑɫɫɑ/, which means "Swearing" in English) (Thackston, 2006).

The phonology for the Kurdish language is as follows. Generally, in Sorani, 9 phonemic vowels exist alongside 26 to 28 consonants, depending on whether pharyngeal sounds are made (/ħ/ and /ʕ/). Table 1 shows the Kurdish Sorani vowels. Nonphonemic vowels are given between parentheses, and they are shown in the table due to their prominence. Thus, the characters in Kurdish Sorani, by relying on where they appear in a word, take different shapes. The vowels (æ) are occasionally pronounced as (ə); this sound is similar to that of the first syllable of English words such as "around". This pronunciation could change if it is followed by sound (j), which is the same as (y) in English, or when (æ) is preceded by (w); however, if it is preceded by (j) and (j) is a part of another syllable, then it is pronounced as (ɛ), similar to the word "bell" in English.

Table 1: Kurdish Sorani Vowels

| IPA | Sorani | Kurmanji | Sorani word examples | English word examples |
|---|---|---|---|---|
| i | ى | î | brin = "wound" | "seat" |
| ɪ | - | i | hemin = "patient or quiet" | "sit" |
| e | ێ | e, ê | rebar = "guide" | "bed" |
| (ɛ) | ە | e | bɛjɑn = "sunrise" | "bet" |
| (ə) | ەا | (mixed) | bərɛz = "respectful" | "but" |
| æ | ە | â | tæll = "bitter" | "cat" |
| u | وو | û | dur = "far" | "moon" |
| ʊ | و | u | gʊrg = "wolf" | "cook" |
| o | ۆ | o | mor = "stamp" | "got" |
| ɑ | ا | a | dɑw = "trap" | "calm" |

Various consonants are expressed as Kurdish Sorani letters, depending on the position of its occurrence in a word. The consonants for Kurdish Sorani are shown in Table 2:

Table 2: Kurdish Sorani Consonants

| IPA | Sorani | Kurmanji | Sorani word examples | English word examples | Annotation |
|---|---|---|---|---|---|
| b | ب | b | باڵ bał (wing) | *b* in "buy" | |
| p | پ | p | پان pan (wide) | *p* in "peek" | |
| t | ت | t | تاج taj (crown) | *t* in "time" | |
| d | د | d | دار dar (tree) | *d* in "deer" | |
| k | ک | k | کۆکە koka (cough) | *c* in "cat" | |
| g | گ | g | گا ga (bull) | *g* in "green" | |
| q | ق | q | قووڵ qûł (deep) | Similar to K in English but sounds deeper in the throat | |
| ʔ | ا | ' | ئازاد âzad (free) | middle sound in "uh-oh" | |
| f | ف | f | فیل fil (elephant) | *f* in "fire" | |
| v | ڤ | v | ڤێلا vela (vela) | *v* in "velocity" | |
| s | س | s | سوور sûr (red) | *s* in "spring" | |
| z | ز | z | زریان zirjan (hurricane) | *z* in "zinc" | |
| x | خ | kh | خراپ khrap (bad) | similar to *ch* in German "Bach" | |
| ʕ | ع | ` | عێراق 'irâq (Iraq) | pharyngeal (like Arabic "ain") | the sound is expressed in Arabic words; it is nonexistent in Kurdish words |
| ɣ | غ | gh | غاز gaz (gas) | similar to ع sound, but voiced | normally pronounced [x] |
| ʃ | ش | sh | شین shin (blue) | *sh* in "shape" | |
| ʒ | ژ | zh | ژیان zhyan (life) | *ge* in "garage" | |
| tʃ | چ | ch | چاک châk (good) | *ch* in "chicken" | |
| dʒ | ج | j | جوان jwân (beautiful) | *j* in "juice" | |
| ħ | ح | ḥ | حوشتر hoshtr (camel) | more guttural than the English *h* | commonly utilized in Iraqi dialects, depending on the region |
| h | ھ | h | ھیوا hiwa (hope) | *h* in "hole" | |
| m | م | m | مار mâr (snake) | *m* in "mine" | |
| n | ن | n | نان nân (bread) | *n* in "north" | |
| w | و | w | وڵات wiłât (country) | *w* in "wave" | |
| j | ی | y | یوسف yusif (josef) | *y* in "yard" | |
| ɾ | ر | r | فریاد firyad (rescuer) | *r* in "lord" | |
| r | ڕ | ř, rr | ڕێوی řewi (fox) | Like Spanish trilled *r* | |
| l | ل | l | لێو lew (lip) | *l* in "light" (front of the mouth) | |
| ɫ | ڵ | ł | ساڵ sâł (year) | *l* in "all" (back of the mouth) | |

Nouns in Sorani can occur in three forms: absolute, indefinite, and definite. Absolute state nouns occur without suffixes, such as those in a vocabulary list or a dictionary. Absolute nouns have a generic understanding, such as "chya barza", "Mountain is high", and "hangwen shirina", "honey is sweet". For indefinite nouns, an interpretation will be accepted as in English nouns, and are preceded by any, some, a, or an. Some modifiers can change indefinite nouns, which are listed below:

- Chî (tʃi) means "what"
- Har (haɾ) means "each"
- Hamu (hamu) means "every"
- Chand (tʃand) means "a few"

In addition, nouns in the indefinite state can have one of the following endings: a vowel for singular (yek) or plural (yân) words or a consonant for singular (ek) or plural (ân) words, as illustrated in Table 3:

Table 3: Indefinite State Nouns and Modifiers

| Endings with | For singular | For plural | Singular example | Plural example |
|---|---|---|---|---|
| vowel | yek | yân | ئێوارە ewara "evening" >> ئێوارەیەک ewara**yek** "an evening" | پەنجەرە panjara "window" >> پەنجەرەیان panjara**yân** "(some) window" |
| consonant | ek | an | ژن zhn "woman" >> ژنێک zhn**ek** "a woman" | پەنجەرە panjara "window" >> پەنجەرەیان panjara**yân** "(some) window" |

The interpretation of indefinite state nouns is similar to that for English nouns, which are anticipated by *the*. The following endings can appear for indefinite state nouns, as shown in Table 4:

Table 4: Definite Nouns Modifiers

| Ending with | For singular | For plural |
|---|---|---|
| A vowel | ka | kân |
| A consonant | aka | akân |

However, if a stem noun ends with (i), then it will be mixed with a definite state suffix, and therefore, (eka) or (i+aka, which leads to (eka)) is pronounced.

Verbs in Sorani have present and past stems. To create a simple present verb, the present stem is preceded by the prefix (da or a) and followed by endings with the personal suffix. Table 5 shows an example for the verb (بینین/bininm means "to see"):

Table 5: Present Stem Verb

| Verb | Meaning in English |
|---|---|
| دەبینم dabinm | I see |
| دەبینی dabini | You see (singular) |
| دەبینێ debine | He/She/It sees |
| دەبینین debinin | We see |
| دەبینن dabinn | You see (plural) |
| دەبینن dabinn | They see |

The personal endings for both the second and third person plural (you and they) are the same. Additionally, for a simple past verb, a past stem verb will be used. Table 6 represents the combination of an intransitive verb (خەوتن khawtn, meaning "to sleep") in the form of the simple past verb, for which the past stem of the word خەوتن is خەوت /khawt:

Table 6: Simple Past Tense Verb in Sorani

| Verb | Meaning |
|---|---|
| خەوتم khawtm | I sleep |
| خەوتی khawti | You sleep (singular) |
| خەوت khawt | He/She/It sleeps |
| خەوتین khawtin | We sleep |
| خەوتن khawtn | You sleep (plural) |
| خەوتن khawtn | They sleep |

In Sorani, the transitive past is distinctive in that the agent affix to express possessive pronouns commonly anticipates the stem verb. Table 7 represents an example of the transitive verb خواردن / xwardin, "to eat", in the past form with the object گۆشت/gosht "meat"; the past form of the verb is " خوارد /xward":

Table 7: Past Stem Verb in Sorani

| Verb | Meaning |
|---|---|
| گۆشتم خوارد gosht-m xward | I ate meat |
| گۆشتت خوارد gosht-t xward | You ate meat (singular) |
| گۆشتی خوارد gosht-i xward | He/She/It ate meat |
| گۆشتمان خوارد gosht- mân xward | We ate meat |
| گۆشتتان خوارد gosht- tân xward | You ate meat (plural) |
| گۆشتیان خوارد gosht- yân xward | They ate meat |

From the example, in Table 7, the clitics attached to the object are understood as possessive pronouns. The integration of gosht**-m** is interpreted as "my meat", "gosht**-t**" as "your meat" and so on for the rest. Thus, the clitic that has to be attached to the earlier word is called the agent affix. When the phrasal verb has words except for the verb itself, as shown in Table 7, in the phrasal verb, it will be added to the first word. Otherwise, if the pre-verb implication does not exist, then it will be added to the first morpheme of the verb. In the case of the past progressive when the stem verb is preceded by "da", then the clitic will be added to "da", as shown in the following example with the verb (nusin/to write):

- da**-m** nusi (**I** was writing)
- da**-t** nusi (**You** were writing)

Regarding gender, Sorani contradicts Kurmanji. In Sorani, there is no gender differentiation. This means that there is no pronoun to differentiate between feminine and masculine.

In this paper, a system has been developed and applied to the Kurdish language for the Sorani and Kurmanji dialects. There were some barriers to constructing this system. First, there is not an adequate Kurdish corpus or Kurdish text for both the Sorani and Kurmanji dialects. Second, while constructing the N-grams, some of the letters (ە، ۆ، ئ، ڵ، ژ، پ) are unreadable in different kinds of software, such as R-studio and Python. In the following section, the N-gram language model is explained in detail.

**4. N-gram Language Model**

Language models assign probabilities to a series of words or a sentence or the probability of the next word given a preceding group of words. These models can be useful in a variety of fields, such as spell correction, speech recognition, machine learning, etc. (Ajitesh Kumar, 2018). The following scenarios illustrate the uses of language models.

*4.1 First Scenario:* The calculation of the probability of word sequences is dependent on the probability of each word. For example, say we want to calculate the occurrence probability for this sentence (بیمەی ئۆتۆمبیل پێویستە بە ووریاییەوە بکردرێت). The sentence's meaning in English is "(Car insurance must be bought carefully.")". Table 8 shows how to pronounce the Kurdish words, alongside the meaning of each word:

Table 8: First Scenario Kurdish Sentence

| Kurdish words | How to pronounce the Kurdish words | Meanings in English |
|---|---|---|
| بیمەی | bimey | insurance |
| ئۆتۆمبیل | otombil | car |
| پێویستە | pewiste | must |
| بە | ba | be |
| ووریاییەوە | wiryayewe | carefully |
| بکردرێت | bikirdret | bought |

Therefore, the calculation is as follows:

$$P\ (\text{بیمەی ئۆتۆمبیل پێویستە بە ووریاییەوە بکردرێت}) = P(\text{بیمەی})P(\text{ئۆتۆمبیل})P(\text{پێویستە})P(\text{بە})P(\text{ووریاییەوە})P(\text{بکردرێت}) \tag{1}$$

Thus, the probability of the sentence (بیمەی ئۆتۆمبیل پێویستە بە ووریاییەوە بکردرێت) is equal to the multiplication of the probability of each word in the sentence, and the probability of every word is equal to the number of words occurring in the corpus divided by the summary of the word in the corpus; it can be computed as follows:

$$P\ (\text{ئۆتۆمبیل}) = \frac{the\ times\ of\ occurence\ (\text{ئۆتۆمبیل})\ in\ the\ text\ corpus}{overall\ number\ of\ words\ in\ the\ text\ corpus} \tag{2}$$

From the above equation, the probability of word (ئۆتۆمبیل) is equal to the number of times the word (ئۆتۆمبیل) occurred in the corpus divided by the total number of words in the text corpus.

If we conduct the above calculation general, then we can say:

$$P(w_i) = \frac{c(w_i)}{c(w)} \tag{3}$$

In the above equation, the representation of each part is as follows:

$w_i$ represents any determined word, c($w_i$) represents the frequency of the determined words, and c($w$) serves as the frequency of all words. Then, the second scenario is illustrated as shown in the following subsection.

*4.2 Second Scenario:* The probability calculation of a sequence of words is dependent on the probability of words in the preceded word occurrence. For instance, say we want to compute the occurrence probability of the sentence (باشترین ویبسایت بۆ بەراورد کردنی نرخی ئۆتۆمبیل). The meaning of this sentence in English is "The best website for comparing car prices"). Table 9 illustrates the Kurdish words in the sentence.

Table 9: Second Scenario Kurdish Sentence

| Kurdish words | How to pronounce the Kurdish words | English meaning |
|---|---|---|
| باشترین | bashtreen | best |
| ویبسایت | website | website |
| بۆ | bo | for |
| بەراوردکردنی | berawirdkirdiny | comparing |
| نرخی | nirkhy | price |
| ئۆتۆمبیل | otombil | car |

Therefore, the probability of occurrence of the above sentence is as follows:

$P$(باشترین ویبسایت بۆ بەراوردکردنی نرخی ئۆتۆمبیل) = $P$(باشترین/ $start\ of\ the\ sentence$)$P$(ویبسایت/باشترین)$P$(بۆ / ویبسایت) ... $P(end\ of\ sentence$ / ئۆتۆمبیل) (4)

From the above computation, the probability of a word given the preceded word can be measured as follows:

$$P(\text{ویبسایت / باشترین}) = \frac{P(\text{ویبسایت باشترین})}{P(\text{باشترین})} \tag{5}$$

From the above equation, the probability of the word ویبسایت given the word باشترین is equal to the probability of the occurrence of both words together (باشترین ویبسایت) divided by the probability of the given word, which is (باشترین).

The above equation, in general, is as follows:

$$P\frac{W_i}{W_{i-1}} = \frac{P(W_{i-1}\ W_i)}{P(W_{i-1})} \tag{6}$$

The probability of a word given the previous word is dependent on the results of the probability of the word and the previous word together divided by the probability of the previous word, which is independent.

Therefore, in the N-gram language model, several features or grams exist, such as the unigram, bigram, and trigram, and so on. In the next subsection, the unigram language model is explained.

*4.3 Unigram Language Model:* Say we want to find the probability of ( باشترین کامه ) (خزمەتگوزاری ئینتەرنێتی هەیە). The meaning of the sentence in English is "which has the best internet services". According to the unigram model, Table 10 shows the pronunciation of Kurdish words with their meanings.

Table 10: Unigram Language Model of The Example Kurdish Sentence

| Kurdish words | How to pronounce the Kurdish words | English meaning |
|---|---|---|
| کامە | kamaa | which |
| باشترین | bashtreen | the best |
| خزمەتگوزاری | khizmetguzary | service |
| ئینتەرنێتی | internet | internet |
| هەیە | heya | has |

The probability measurement is as follows:

$$P(\text{کامە باشترین خزمەتگوزاری ئێتەرنێتی هەیە}) = P(\text{کامە})P(\text{باشترین})\ldots P(\text{ئینتەرنێتی})P(\text{هەیە}) \tag{7}$$

From the above measurement, each word occurrence probability from the text corpus for every word ($w_i$) is calculated below:

$$P(w_i) = \frac{c(w_i)}{c(w)} \tag{8}$$

where $w_i$ is the *ith* word, $c(w_i)$ is the frequency of $w_i$, and $c(w)$ is the frequency of a total of the words. The bigram language model is explained next.

*4.4 Bigram Language Model:* If we use the previous example with the bigram model, the probability can be computed as follows:

$$P(\text{کامە باشترین خزمەتگوزاری ئینتەرنێتی هەیە}) = P\left(\frac{\text{کامە}}{start\ of\ sentence}\right) P\left(\frac{\text{باشترین}}{\text{کامە}}\right) P\left(\frac{\text{خزمەتگوزاری}}{\text{باشترین}}\right) \ldots P(\frac{end\ of\ sentence}{\text{هەیە}}) \quad (9)$$

From the above equation, the probability of the sentence depends on and is equal to the probability of the first word in the sentence. Then, the probability of the next word over the previous word until it reaches the last word of the sentence and each of the probabilities is computed as follows:

$$P\left(\frac{\text{خزمەتگوزاری}}{\text{باشترین}}\right) = P\left(\frac{\text{خزمەتگوزاری باشترین}}{\text{باشترین}}\right) \quad (10)$$

Therefore, the probability of word (خزمەتگوزاری) over the probability of the word (باشترین) is equal and depends on the result of the probability of both of the words divided by the probability of the word باشترین. In another way, the above calculation can be conducted as follows:

$$P(\frac{\text{خزمەتگوزاری}}{\text{باشترین}}) = \frac{c(\text{باشترین خزمەتگوزاری})}{c(\text{باشترین})} \quad (11)$$

Explanation of the above equations: the probability of word (خزمەتگوزاری) given word (باشترین) is equal to the probability of word (باشترین خزمەتگوزاری) divided by the probability of word (باشترین). Alternately, the probability of word (ئینتەرنێت) given word (باشترین) is equal to the division of frequency of the word (باشترین ئینتەرنێت) by the frequency of the word (باشترین). In general, the above expression can be addressed as the probability of each word given the preceded or previous word. Consequently, the probability of each word given an earlier word is represented as $\frac{w_i}{w_{i-1}}$, and it can be computed as follows:

$$P\frac{w_i}{w_{i-1}} = \frac{P(w_{i-1}\ w_i)}{P(w_{i-1})} \quad (12)$$

This equation is the general form of calculating the bigram model, which depends on the probability of giving two previous words ($w_i$, $w_{i-1}$).

In the following subsection, the trigram language model is discussed.

*4.5 Trigram Language Model:* Say we want to calculate the probability of (کام کۆماپنیایە باشترین خزمەتگوزاری ئینتەرنێت دابین دەکات). The English meaning of this sentence is "which company provides the best internet services". Table 11 shows how to pronounce the Kurdish words for the example using a trigram language model.

Table 11: Trigram Language Model for a Kurdish Sentence Example with the Meaning and Pronunciation of Each Word

| Kurdish words | Way to pronounce the Kurdish words | English meaning |
|---|---|---|
| کام | kam | which |
| کۆمپانیایە | kompanyaye | company |
| باشترین | bashtreen | the best |
| خزمەتگوزاری | khizmetguzary | service |
| ئینتەرنێت | internet | internet |
| دابین دەکات | dabin | provides |

Then, the calculation of the probability using the trigram model is as follows:

$$P(\text{کام کۆمپانیایە باشترین خزمەتگوزاری ئینتەرنێت دابین دەکات}) = P\left(\frac{\text{کۆمپانیایە}}{\text{کام is start of sentence}}\right) P\left(\frac{\text{باشترین خزمەتگوزاری}}{\text{باشترین}}\right) P\left(\frac{\text{ئینتەرنێت}}{\text{باشترین خزمەتگوزاری}}\right) \ldots P\left(\frac{\text{end of sentence}}{\text{دابین دەکات}}\right) \quad (13)$$

The calculation of the probabilities is represented below:

$$P\left(\frac{\text{باشترین}}{\text{کام کۆمپانیایە}}\right) = P\left(\frac{\text{کام کۆمپانیایە باشترین}}{\text{کام کۆمپانیایە}}\right) \quad (14)$$

In another way, it can be computed as follows:

$$P\left(\frac{\text{باشترین}}{\text{کام کۆمپانیایە}}\right) = \frac{c(\text{کام کۆمپانیایە باشترین})}{c(\text{کام کۆمپانیایە})} \quad (15)$$

The above calculations in equations 14 and 15 can be explained as the probability of word (باشترین) given words (کام کۆمپانیایە) being equal to the probability of word (کام کۆمپانیایە باشترین) divided by the probability of words (کام کۆمپانیایە). Likewise, the probability of word (باشترین) given words (کام کۆمپانیایە) is equal to the division of the frequency of words (کام کۆمپانیایە باشترین) by the frequency of words (کام کۆمپانیایە).

Overall, the probability of any word given two antecedent words can be expressed as

$\frac{w_i}{w_{i-2}, w_{i-1}}$ , and the computation can be achieved as follows:

$$P(\frac{w_i}{w_{i-2}, w_{i-1}}) = \frac{P(w_{i-2}, w_{i-1}, w_i)}{P(w_{i-2}, w_{i-1})} \qquad (16)$$

The previous text or words ($T_n$) are predicted with a given corpus ($T_1, T_2, ..., T_{n-1}$), which conjectures the probability function $P(T_n | T_1 ... T_{n-1})$. This is the Bayesian theory.

$$P(T) = \sum_{i=1}^{n} P(T_i | T_1 ... T_{i-1}) \qquad (17)$$

A sequence of N-grams can be decreased by using the conditional probability in the Markov Chain rule.

$$P(T_1, T_2, ..., T_n) = P(T_1) P(T_2 | T_1) ... P(T_n | T_1, ... T_{n-1}) \qquad (18)$$

The expression of the N-gram model is $P(T_i|T_1 ... T_{i-1}) >> P(T_i|T_{i-n+1}...T_{i-1})$. As a result, if $n=1$, it means (unigram) when $n=2$ means bigram when $n=3$ (trigram) and so on. From a training corpus, the probabilities will be derived, which leads to the construction of the model. The executed N-gram suggestion algorithm speculates that the next word could be predicted depending on the *n-1* preceded word. In the system, when a user types some letters, it gives the correction and the suggested words to the user. The text corpus plays a vitally important role in the system.

In summary, the N-gram language model adjoins a series of *n* items from a provided corpus. In other words, it is the probability of word sequences. The words can be 2 words, 3 words, and so on until *n*-words. Thus, the N-gram language model, briefly, is a sequence of n-words. Consequently, the model is utilized to suggest that the next words depend on a given text corpus. It is a probabilistic model because it is based on probability (Daniel Jurafsky & James H. Martin, 2018). Thus, the preceding word is anticipated in the structure of (*n-1*), which is constructed from the Markov Chain model. The main advantages of the N-gram model are its simplicity and scalability with a greater amount or value of *n* (Daniel Jurafsky & James H. Martin, 2018). As a result, the N-gram can be expressed in Table 12:

Table 12: N-grams Mapping to Word(s)

| N-gram | Word(s) |
|---|---|
| unigram | one word |
| bigram | two words |
| trigram | three words |
| and so on… | |

The Stupid Backoff algorithm is explained in the next subsection.

*4.6 Stupid BackOff:* If texts from the web are utilized, massive language models can be built, as Google did in 2006. They produced a huge set of N-grams, from 1-grams to 5-grams, consisting of five sequences of words utilizing approximately 13 million unique word types (Franz and Brants, 2006). Efficiency is vitally important when forming a language model for large N-grams instead of saving every single word as a string. Additionally, probabilities in general are quantized by utilizing a 4-8 bit rather than an 8-byte float.

N-grams can be dwindled by pruning, for instance, saving N-grams with a greater value of some threshold. In another way, trivial N-grams are pruned by using entropy (Stolcke, 1998). Last, an efficient language model, such as KenLM (Heafield 2011, Heafield et al. 2013), uses stored arrays, so the probability and the backoff are combined efficiently in a single value. Brants et al. (2007) revealed that for a large language model, a simple algorithm might be satisfactory or adequate. Accordingly, the algorithm is known as the Stupid Backoff.

In the Stupid Backoff algorithm, the attempt to build language models by distributing true probabilities is stopped. Thus, the higher-order probability is not discounted. Accordingly, if the higher-order N-gram is counted as zero, then we will backoff to the lower-order N-gram. As a result, a probability distribution is not generated by the Stupid Backoff algorithm (Brants et al. 2007). Therefore, it is referred to $S$ in the following equation:

$$S(w_i \mid w_{i-k+1}^{i-1}) = \begin{cases} \dfrac{count(w_{i-k+1}^i)}{count(w_{i-k+1}^i)} & if\ count(w_{i-k+1}^i) > 0 \\ \lambda S\ (w_i \mid w_{i-k+1}^{i-1}) otherwise \end{cases} \quad (19)$$

The backoff in the algorithm will be stopped in the unigram, and the precomputed and saved frequency of the word is represented as $S$ and is equal to $(w) = \dfrac{Count(w)}{N}$, $N$ is the size of the corpus, and $\lambda$ is the weight of the backoff. Additionally, it is discovered that the value of $\lambda$ will be perfect when it is equal to 0.4, and the backoff variable is made to rely on $k$.

## 5. Proposed System

In this section, the methodology to predict the next words for Kurdish Sorani and Kurmanji dialects is described.

**RStudio:** RStudio is a free and open-source software integrated development environment (IDE) that allows users to engage with R programming easily. R must be installed before RStudio. The R programming language is utilized for statistical computing and graphics. Therefore, in RStudio, the interface is well formed due to source codes, tables, and graphs that can be distinctly viewed by users. In addition, RStudio allows users to import several kinds of files, such as CSV, Excel, and SAS files, into RStudio (Oscar, 2013). RStudio is available for all the platform windows, MacOS, and Linux. Figure 2 shows the RStudio interface:

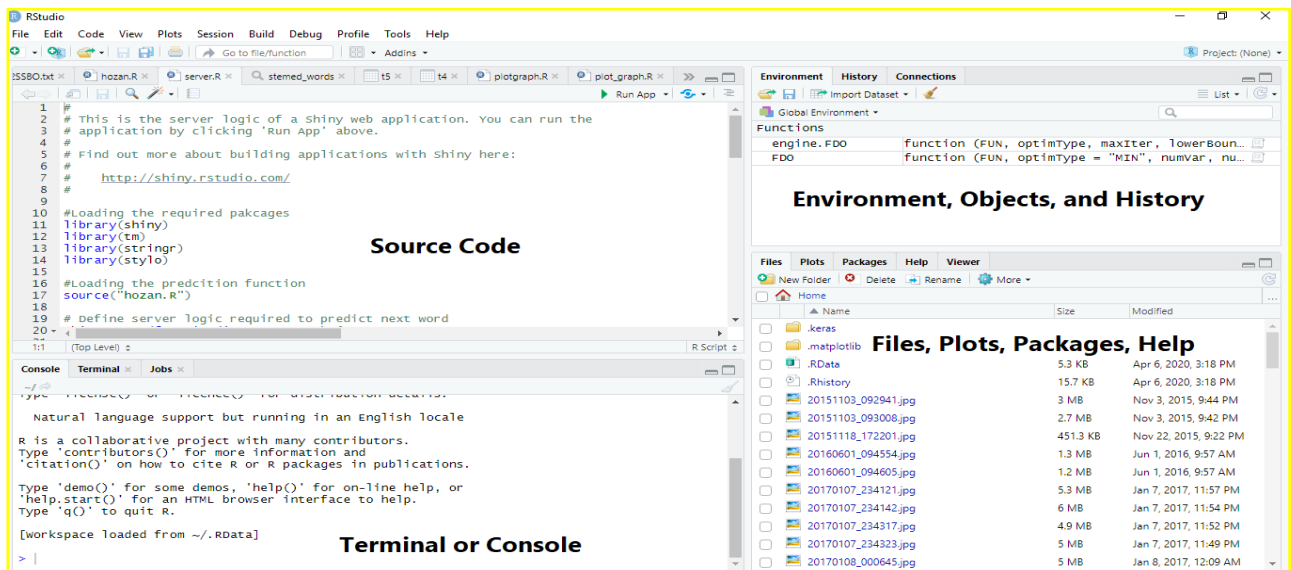

Figure 2: RStudio Interface

Figure 3 illustrates the phases of the proposed system:

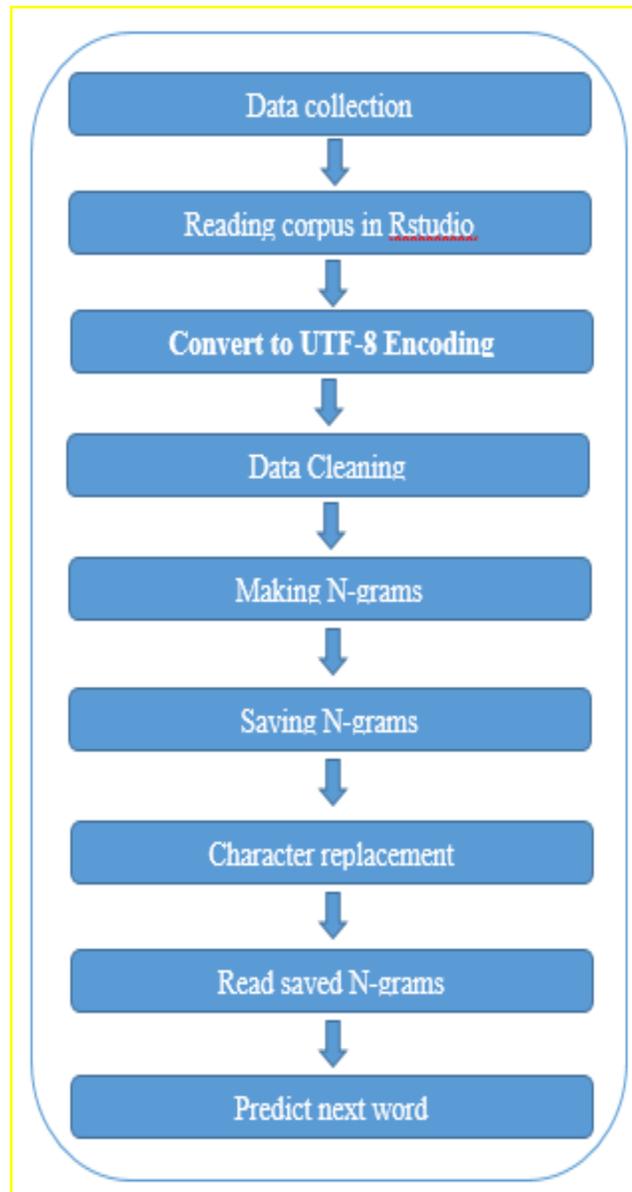

Figure 3: Methodology of Next Word Prediction for Kurdish Sorani and Kurmanji

Figure 4 shows the details of the proposed system.

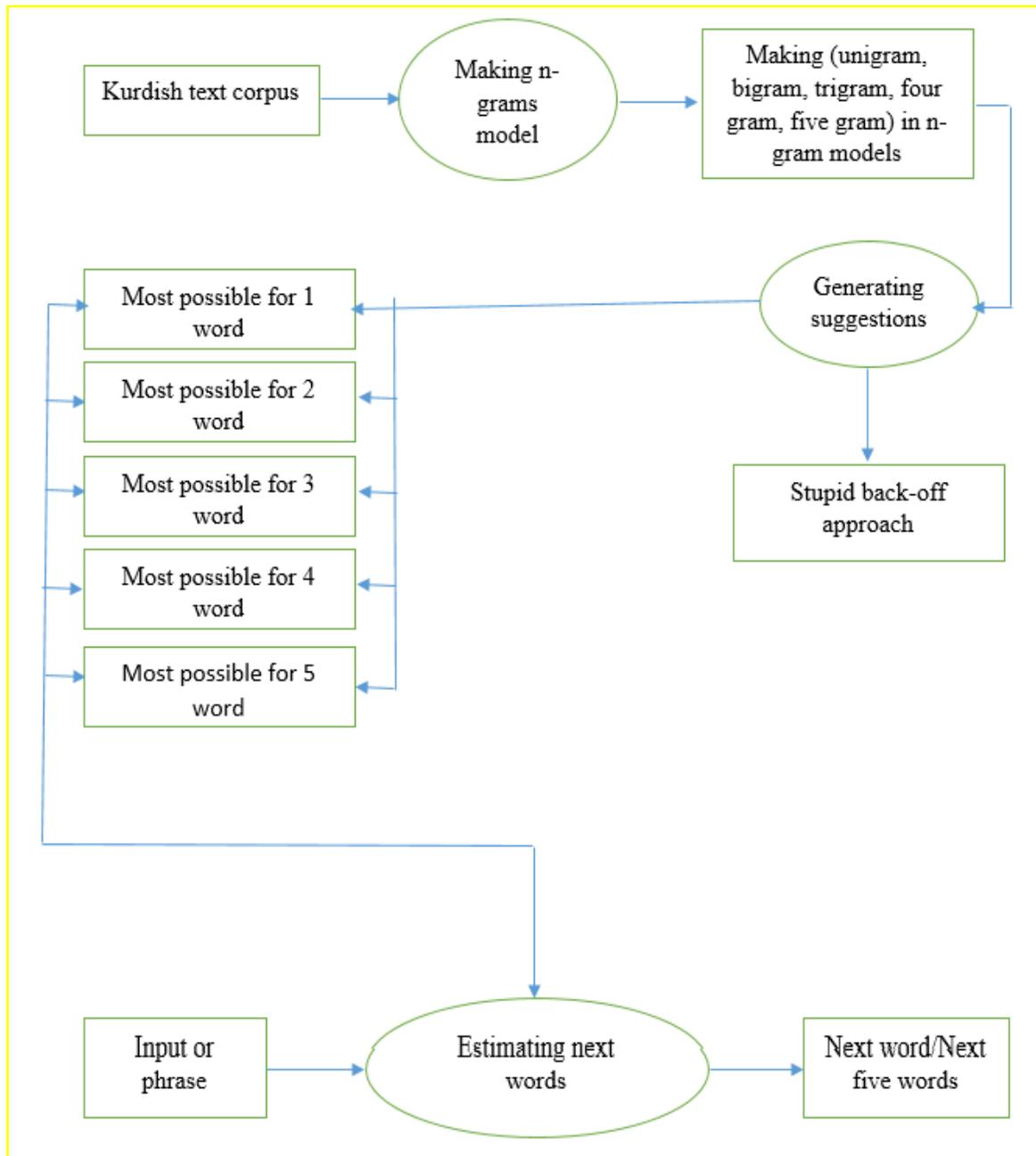

Figure 4: Proposed System

***5.1 Data Collection:*** A Kurdish text corpus is amassed from different sources, such as websites and PDF books. The corpus contains more than 500,000 words and includes words of different topics, such as social, religious, economic, etc. Kurdish Sorani texts are different from Kurmanji text, for instance, in terms of the morphology of the letters. Accordingly, all the texts are transformed into Kurmanji using the Kurdish Sorani (Kurmanji) converter. An example is shown in Figure 5:

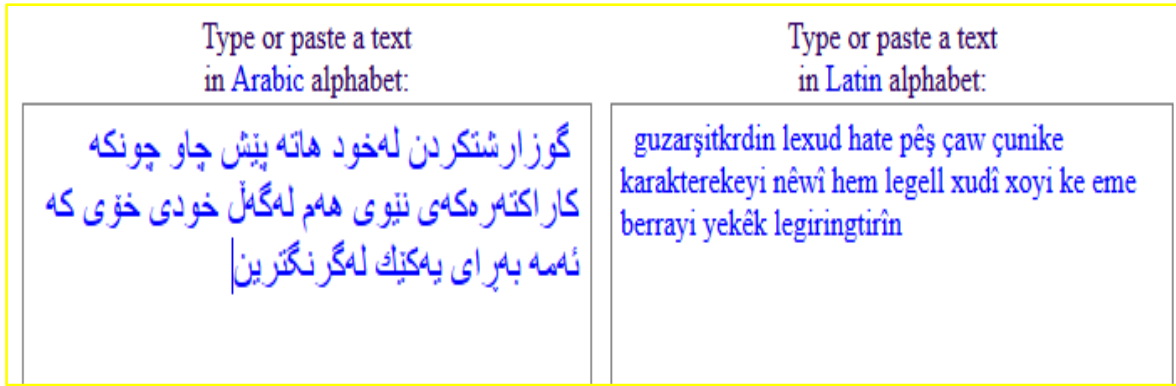

Figure 5: Kurdish Sorani to Kurmani (Latini) Converter

All Kurdish texts obtained from different sources are merged into one text file for Sorani, and then all the texts are transformed into Kurmanji. Table 13 shows the sources of the data collection.

Table 13: Data Collection Resources

| sources | links |
|---|---|
| **website** | http://www.rudaw.net |
| **website** | http://www.nrttv.com |
| **website** | http://www.knnc.net |
| **website** | https://www.kurdistan24.net/ |
| **pdf book (gamzha)** | http://ktebstan.com/bookdetails.php?bookID=757 |
| **pdf book (کچەی پرتەقاڵ)** | http://ktebstan.com/bookdetails.php?bookID=676 |
| **pdf book (دەریاوانە خنکاوەکە)** | https://www.kurdipedia.org/default.aspx?q=2011061013315860261&lng=5 |

The text corpus information is shown in Table 14.

Table 14: Kurdish Text Corpus Information

| Size | 860KB |
|---|---|
| **number of words** | 505000 words |
| **link to obtain data** | https://ufile.io/r0mw41sa |

*5.2 Read the Corpus on RStudio:* By using a function in Rstudio, the processed corpus is read by the software. This corpus is then converted into a vector term document. As a result, the corpus is ready to use to construct N-grams, as shown in the code snippet below:

> *Kurdish_data <- convert_text(Kurdish_data)*

*5.3 Convert to UTF-8 Encoding:* In this phase, the collected texts are encoded to UTF-8 for Kurdish Sorani, but for Kurdish Kurmaji, this step is not necessary. Thus, the corpus text is read on Rstudio and then encoded to UTF-8, as shown in the following code snippet:

> *Kurdish_data <- readLines("kurdishsoranicorpus.txt", encoding = "UTF-8", warn = FALSE)*

*5.4 Data Cleaning:* In this step, a function in Rstudio is used to eliminate punctuation marks, double quotation, and numbers, for instance, 1, 0.2, ", ', ?, !, ;, :, ., etc. In addition, for Kurdish Sorani, non-Kurdish words or letters are omitted. For this purpose, the gsub function is used in Rstudio, as shown in the following code snippet:

> *text_Kurdish <- gsub("[^[:alpha:][:space:][:punct:]]", "", text_Kurdish);*
> 
> *text_Kurdish<-removePunctuation(text_Kurdish)*

The same process is applied for numbers, whitespaces, etc.

*5.5 Making N-grams:* At this point, the read text file, which includes the corpus, is used alongside a series of actions to created N-grams, such as N-gram1, N-gram2, and so on. These N-grams are created as objects inside the software. Therefore, a function is utilized to build N-gram tokens during the term-document matrix construction. For this purpose, to achieve this goal, a library called Weka is used. To create each N-gram, the min and max value are determined and the numbers of word or words are determined. For example, to make a unigram, the min and max numbers have to be equal to one, as expressed in the code snippet below:

> *unigram <- function(x) NGramTokenizer(x, Weka_control(min = 1, max = 1))*

Then, the term document matrices will be built by using the following code snippet:

> *unigram <- TermDocumentMatrix(Kurdish_data, control = list(bounds=list(global = c(2, Inf))))*

The others, including a bigram, which has 2 grams, and a trigram, which has 3 grams, are achieved by repeating the above step, but we have to change the min and max to 2 for the bigram and to 3 for the trigram and so on for the rest. Accordingly, the functions will be executed as follows:

> *bigram <- function(x) NGramTokenizer(x, Weka_control(min = 2, max = 2))*
>
> *trigram <- function(x) NGramTokenizer(x, Weka_control(min = 3, max = 3))*

*5.6 Saving* **N-grams:** In this phase, the built N-grams are saved in Excel files by using the R function in RStudio. To do so, a library called xlsx has to be installed and used. Then, the Write.xlsx function is used to export the created N-gram objects to an Excel file. After that, the object name is written as the name of the file followed by the.xlsx extension. This file name should be between two double quotations, as illustrated below:

> *write.xlsx(ngram1, "unigram.xlsx")*

The saved file consists of two columns, word and freq, which is the abbreviation for frequency.

After exporting, a challenge is faced for the Kurdish Sorani dialect, as several Kurdish letters cannot be read in Rstudio objects, such as (ە، ۆ، ێ، ڵ، ژ). Each of the characters has a UTF-8 encoded code or symbol; for example, (ە) is stored as (U+0647), and each of the other letters has its own encoding code or symbol. As a consequence, Kurdish letters are inserted in place of these symbols. The details of this step are explained next.

*5.7 Character Replacement:* Each N-gram file is exported to an Excel file. As mentioned in the previous phase, each of the files contains the previously determined Kurdish letters, which cannot be read inside objects. Therefore, all the mentioned encoded characters are manually replaced with the Kurdish Sorani letters. To accomplish this procedure, each file is opened, and by using the replace function inside Excel, the symbols are searched for and replaced by its related Kurdish character. This is repeated for every N-gram, which is in an Excel file.

*5.8 Read Saved Excel Files:* At this point, the saved Excel files or N-grams are read into R-Studio. Namely, all the N-gram files, which are saved as Excel files, are read like a list of Excel files. These files include replacement characters, as shown in the following code snippet:

```
allFrequencies = list(
read.xlsx ("1-gram.xlsx",sheetName = "Sheet1",encoding = "UTF-8"),
read.xlsx ("2-gram.xlsx",sheetName = "Sheet1",encoding = "UTF-8"),
read.xlsx ("3-gram.xlsx",sheetName = "Sheet1",encoding = "UTF-8"),
read.xlsx ("4-gram.xlsx",sheetName = "Sheet1",encoding = "UTF-8"),
read.xlsx ("5-gram.xlsx",sheetName = "Sheet1",encoding = "UTF-8"))
```

At the time of reading the Excel file, the sheet name in the Excel file and the encoding should be determined.

*5.9 Predict The Next Words*: Finally, based on the N-grams, the system suggests the next words; that is, when a user types a letter or a word, it suggests the complete word and potential following words.

The Stupid BackOff algorithm is used, which utilizes a few scenarios at the same time. This means discovering probability with a large number of words, and if adequate evidence cannot be obtained for a word chain, one word will be removed from the chain. For instance, when starting with the tri-gram and there is not sufficient evidence, then the system backs off to the bigram or the unigram. Interpolation means that all the N-grams are mixed to obtain the correct probability value, as shown in the following equation:

$$P(x_i \mid x_{i-2}, x_{i-1}) = \begin{cases} (1 - d(x_{i-2}, x_{i-1}))P(x_{i-2}, x_{i-1}) & if\ count(x_{i-2}, x_{i-1}) > 0 \\ \alpha(x_{i-2}, x_{i-1})\ P(x_i \mid x_{i-1}) & otherwise \end{cases} \quad (20)$$

To address the step-by-step explanation of how the suggestions are made in detail, we look at the following examples, as represented in Table 15 below:

Table 15: Examples of N-grams in Kurdish Alongside Their Meaning in English.

| Kurdish words | Meaning in English | N-gram |
| --- | --- | --- |
| من رۆیشتم | I went | 2-gram |
| من کتێب دەخوێنمەوە | I read books | 3-gram |
| نالی شاعیرێکی گەورەی کوردە | Nali is a great Kurdish poet | 4-gram |

Now, the first of the above N-grams likely appears the most frequently. If we translate these words to English, they mean, "I went". In contrast, the third example or N-gram might not appear very frequently. Translating the third N-gram to English, it means "Nali is a great Kurdish poet". Simply, the third example is an instance of an N-gram that does not appear as frequently as examples 1 and 2 in sentences. It is essential to assign the occurrence probability of an N-gram or the next word occurrence probability in a sequence or series of words.

First, this system could be useful in making decisions, where N-grams can be used together to shape an individual entity; for example, (دوا ناوەندی) means "high school" in English and can be considered as a single word. This approach could also be useful in suggesting the next words. If we have a part of a sentence, for instance, "you can write the answers either in", which is (دەتوانیت وەڵامەکان بنوسیت یان) in Kurdish, the next word is more likely to be "pen" (پێنوس) or "pencil" (خامە) than "university" (زانکۆ).

The system is also effective in correcting spelling mistakes; for example, the correction for "you have" could be "you have". In Kurdish, " پێویسته هبێت" could be corrected as " پێویسته هەبێت", given that the word "have" has the highest occurrence probability after the word

"you". It can be seen that the occurrence probability or frequency of the word, that is, the frequency of the word, is crucial.

Depending on the concept, predictions can be made with the N-gram model. Fundamentally, the prediction of the occurrence of a word in an N-gram model relies on the occurrence of its earlier word, which means *N-1*. How should we go back to the history of word sequences to estimate the next word? If we take the bigram as an example, then the word occurrence prediction uses merely its preceding word (*N-1 =1*). Correspondingly, for a trigram, the prediction of word occurrence relies on the two preceding words, where *N-1 = 2*. Now, we have to understand the approach to achieving a word probability assignment in a series of word occurrences. We must remember that this approach will be applied to collected Kurdish texts or words, which is known as the Kurdish corpus. We can take several sentences as an example instead of the corpus because the corpus is extremely large. However, we consider the following sentences as a corpus, and they are shown in Table 16 with their meanings.

Table 16: Several Examples are Taken as a Corpus

| Kurdish words | Meaning in English |
|---|---|
| ئەو ووتی سوپاست ئەکەم | He said thank you |
| ئەوان چوون بۆ سەیران | They went to a picnic |
| ئەو ئەڕوات بۆ مارکێت | He is going to market |
| کەش و هەوا لێرە خۆشە | Weather is nice here |
| ئەوان ڕۆیشتن بۆ مارکێت بۆ کڕینی پێداویستیەکانیان | They went to the market to buy necessary goods |

Let us take the bigram as an example for the explanation. We can discover the word probability depending only on its preceding word. Therefore, generally, we can argue that the probability of a word is dependent on and equal to (the frequency of the preceded word occurrence "$w_f$" before the word"$w$")/(the summation of times of the earlier word frequency of occurrence inside the Kurdish corpus" $w_f$"). This argument can be expressed as follows:

$$P(word) = \left(count(w_f\ w)\right) / \left(count(w_f)\right) \qquad (21)$$

Applying this model to an example (سوپاست ئەکەم) in English (Thank you), we do the following:

The probability of the word "ئەکەم" followed by the word "سوپاست" is determined and can be written as P(سوپاست | ئەکەم), which is a conditional probability. As a result, the equation is equal to the following:

$P$(سوپاست ئەکەم)=(the number of times "سوپاست ئەکەم" occurs)/(the number of times "سوپاست" occurs)

$P$(سوپاست ئەکەم)= 1 / 1

$P$(سوپاست ئەکەم) =1

We can be sure that when the word "سوپاست" occurs, it will be followed by the word "ئەکەم" because, in our five-sentence corpus example, the words "سوپاست ئەکەم" occurred only once. Accordingly, what will happen if we find that the probability of the preceding words are different in the context of the corpus? If we compute the probability of the word "بازار" (in English, "bazaar"), which comes after the word "بۆ" (in English, "to"), we can say P (بازار | بۆ). Explicitly, we explore the next word's probability, which will be "بازار"given the word "بۆ" and can be achieved as follows:

= (number of times بۆ بازار occurred) / (number of times بۆ occurred)

= 2 / 3= 0.67

This result occurs because in the corpus, one of the words is preceded by the word "سەیران" in English, which means "picnic". As a consequence, P(بۆ | سەیران) = 1 / 3. In the text corpus, only "سەیران" and "بازار" occurred after the word "بۆ", and their probabilities are 1 / 3 and 2 / 3, respectively. Last, with the built system and the mentioned corpus, if the user types the word "بۆ",the system provides two suggestions: "بازار" and "سەیران". This means that the word "بازار" was highly ranked and the word "سەیران" was lowly ranked. This process is continually applied for the rest of the N-grams.

## 6. Results and Discussion

In the proposed system for Kurdish Sorani and Kurmanji, unigrams, bigrams, trigrams, four-grams, and five-grams have been made. The following graphs represent the most frequent words for each of the created N-grams for Kurdish Sorani. The unigram's most frequent words for Kurdish Sorani are represented in Figure 6:

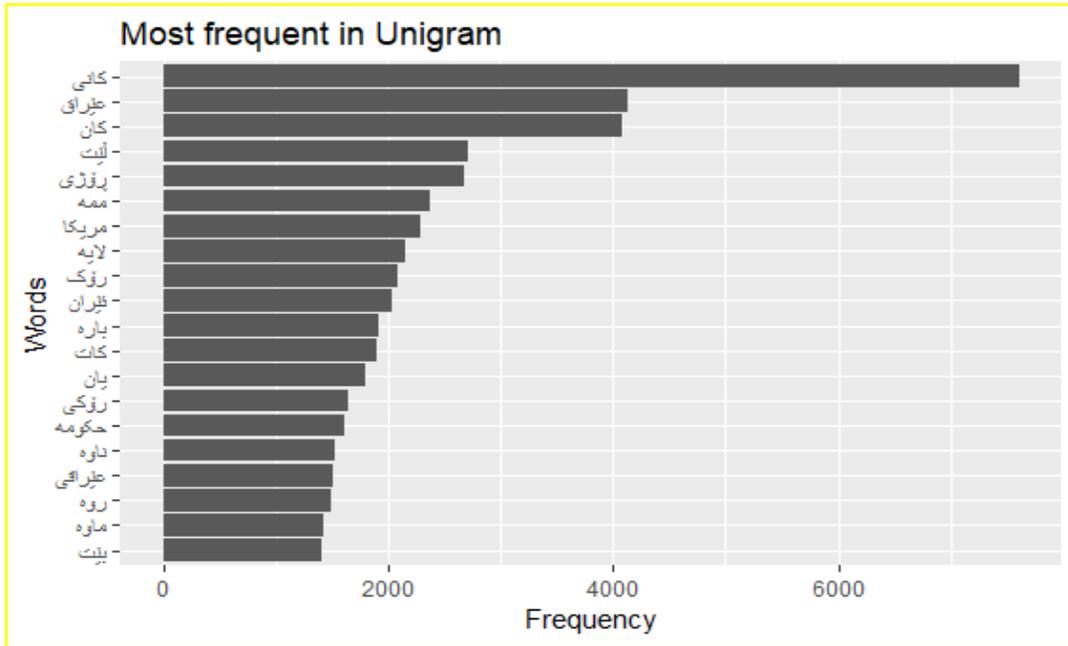

Figure 6: Most Frequent Words in the Unigram for Kurdish Sorani

The most frequent words in the bigram for Kurdish Sorani are shown in Figure 7:

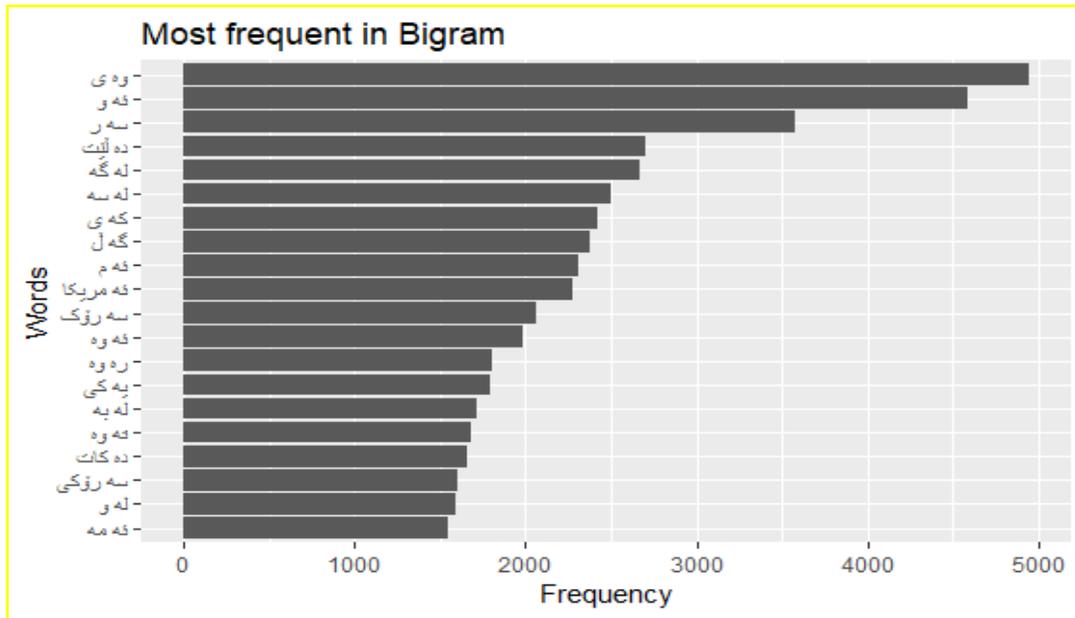

Figure 7: Most Frequent Words in the Bigram for Kurdish Sorani

The top words with the highest frequency in the trigram for Kurdish Sorani are expressed in Figure 8:

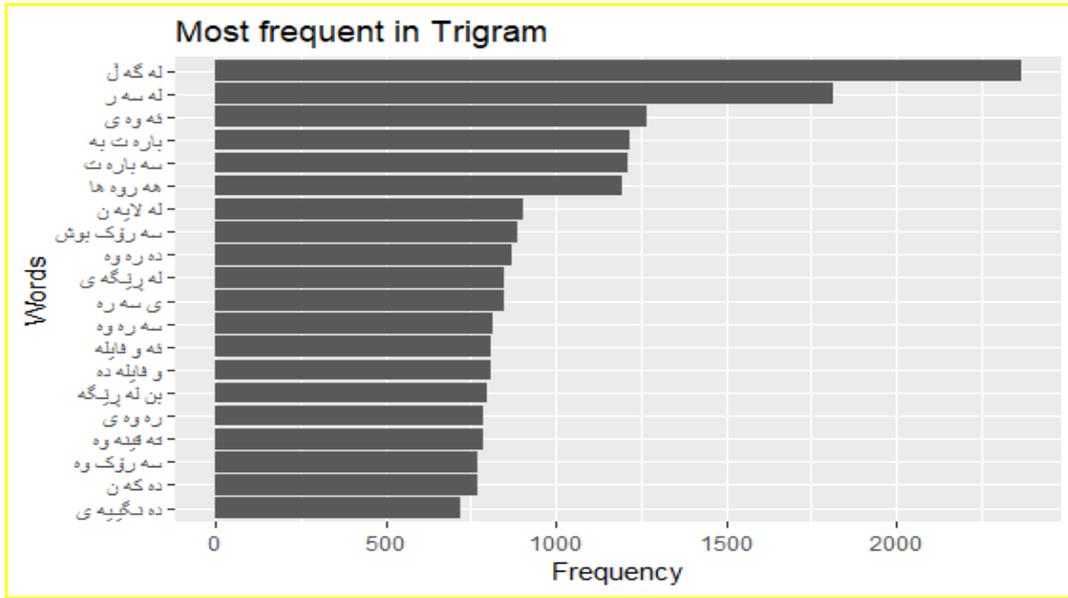

Figure 8: Most Frequent Words in the Trigram for Kurdish Sorani

The highest frequency for words in the four-gram for Kurdish Sorani is given below:

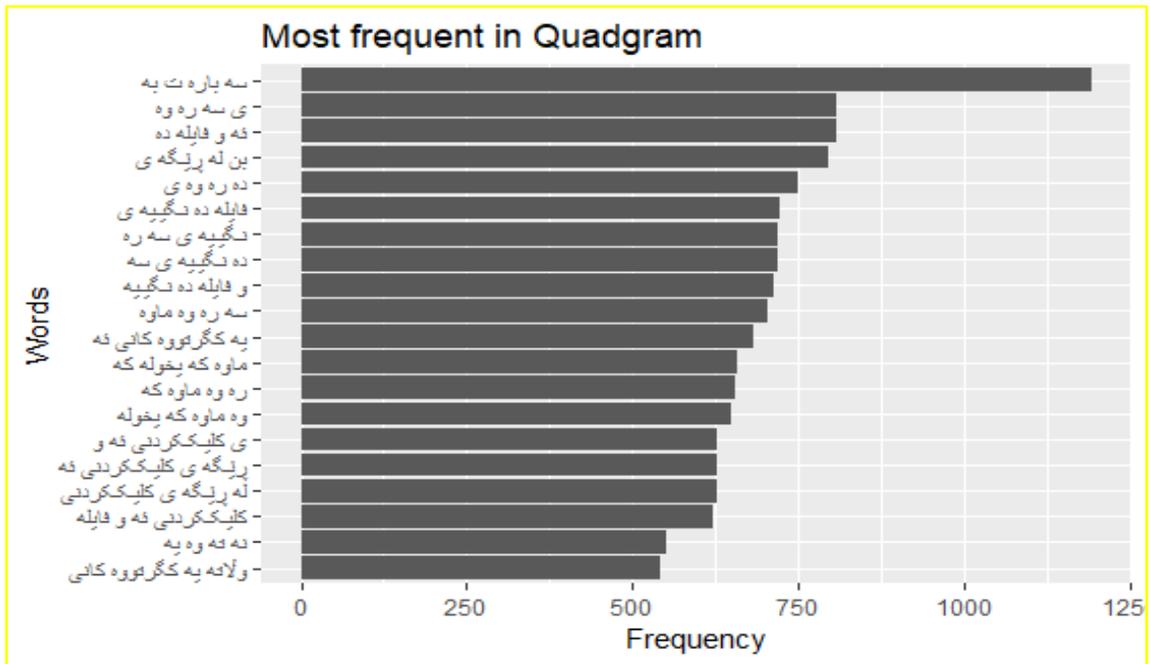

Figure 9: Most Frequent Words in the Four-Gram for Kurdish Sorani

Finally, most frequent words in the five-gram are shown in Figure 10:

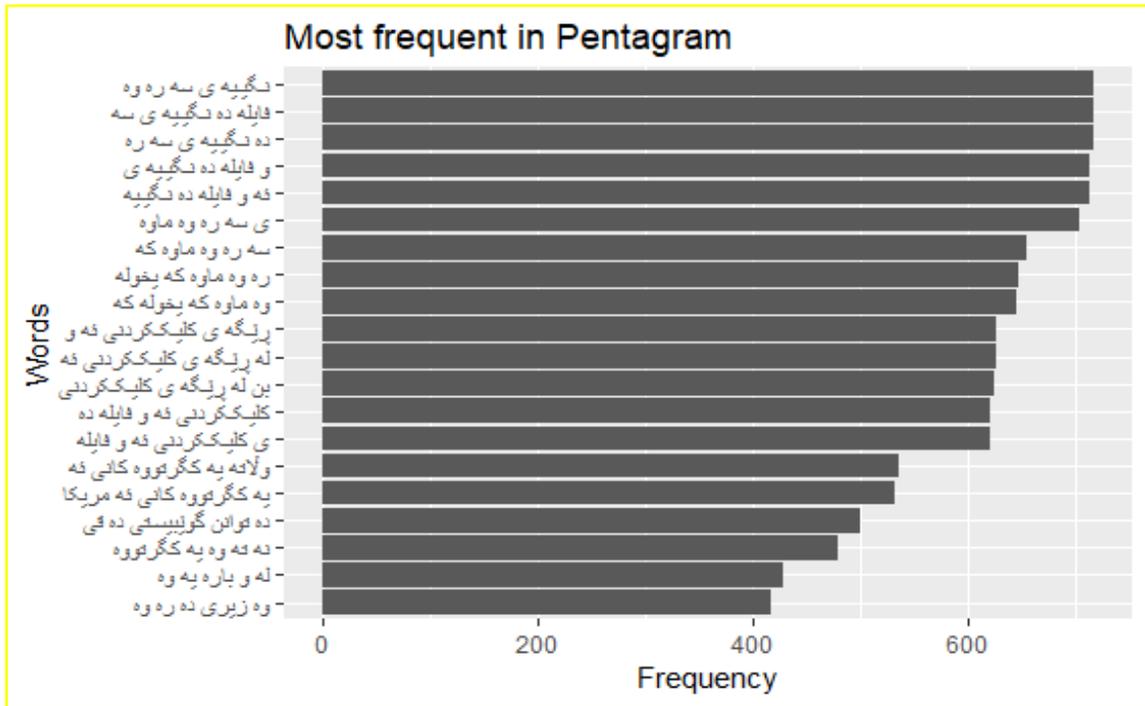

Figure 10: Most Frequent Words in the Five-Gram for Kurdish Sorani

To make predictions, the proposed system takes the input text and then clears and constructs the earlier $n = 1$ to $n-1$. After that, it utilizes a Stupid BackOff algorithm that, combined with weighted value, builds a list of likely next words. For every $n$, depending on the N-gram frequencies, the highest frequency for the top 5 words increases or is matched. Later, it combines the consequences. Then, it sorts them from the highest to lowest likelihood. Last, the top predicted words are shown to the user. The Shiny application was used to produce the interfaces of the systems for Sorani and Kurmanji as follows:

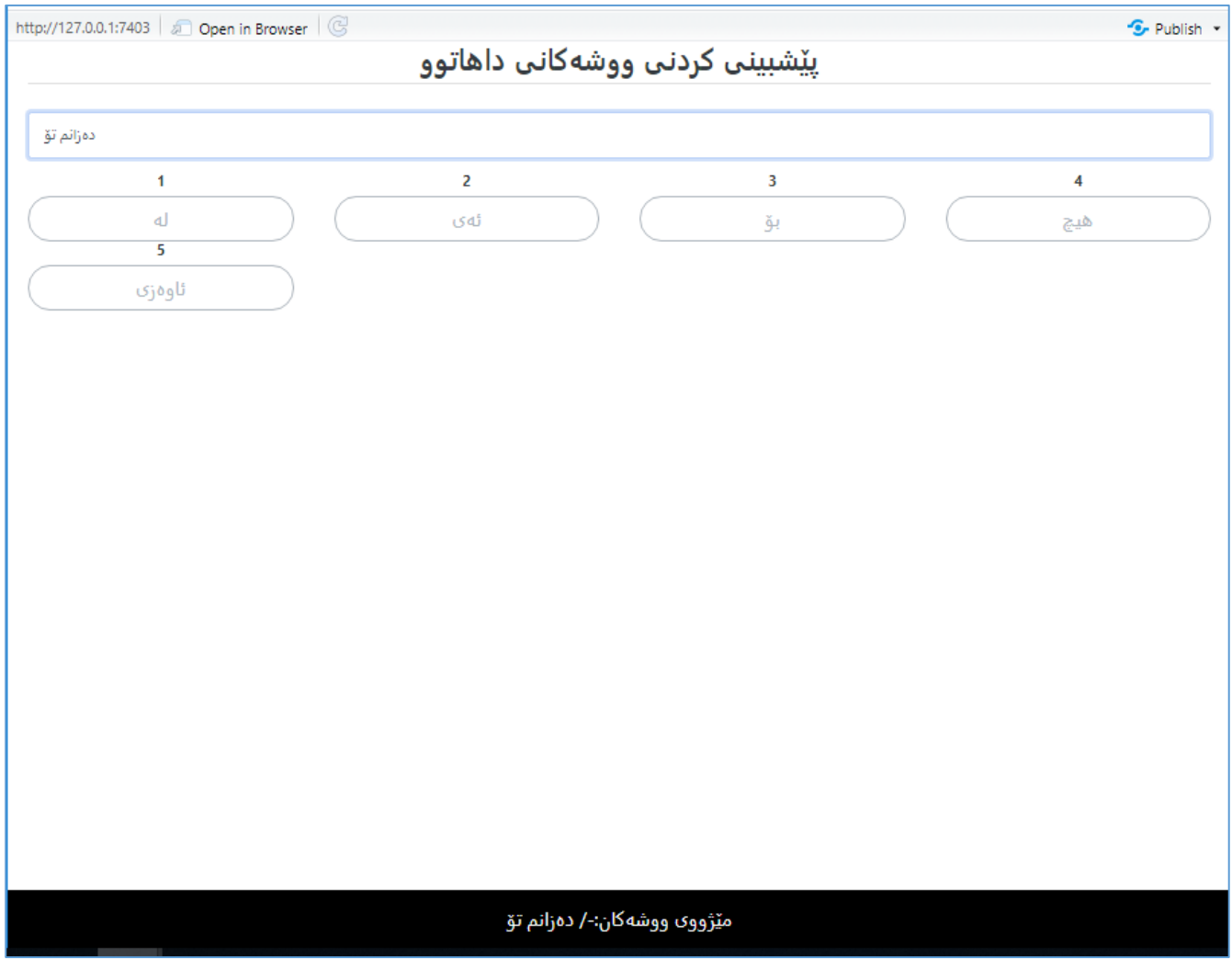

Figure 11: Next Word Prediction for Kurdish Sorani

As shown in Figure 11, a user types a word, and then, the application suggests five words to be selected by the user. The user can input another word, and then, the system predicts the next word by giving five words that can be chosen.

The translations of the above results are shown in Table 17:

Table 17: Translation of the Suggestion Results from Kurdish Sorani to English

| Words in English | Words in Kurdish Sorani |
|---|---|
| I know | دەزانم |
| that | ئەو |
| and | و |
| you | تۆ |
| Kurdish | کوردی |
| this | ئەم |

Additionally, for Kurdish Kurmanji, the Shiny application was used to design the interface, as shown in Figure 12:

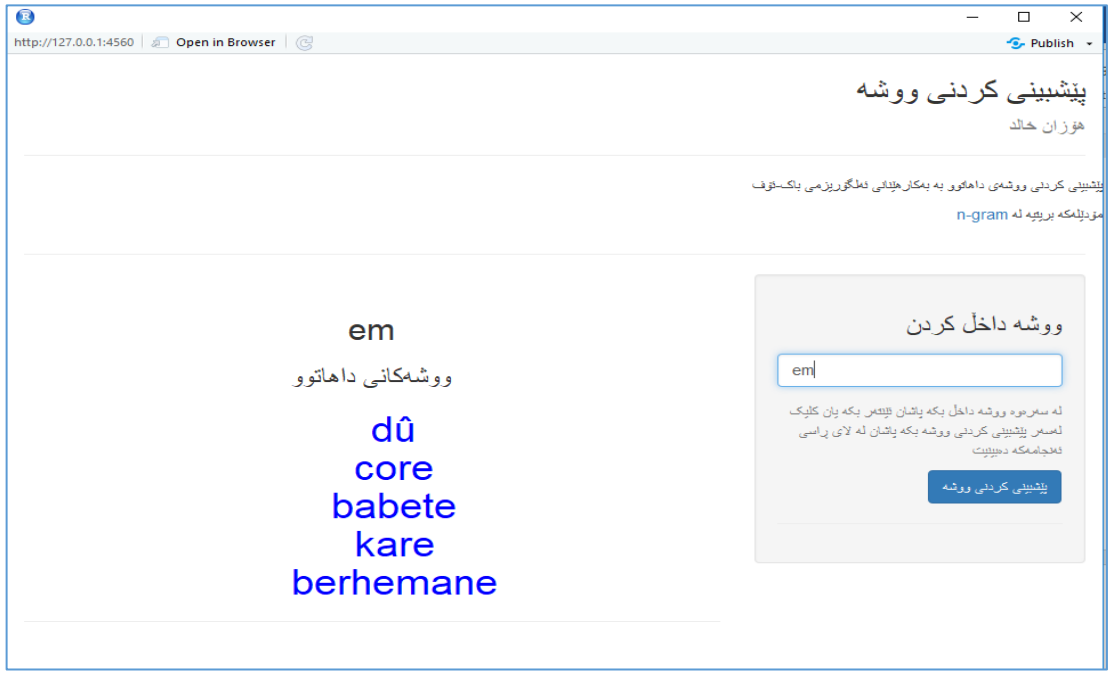

Figure 12: Next Word Prediction for Kurdish Kurmanji

The same process can be done for the Kurdish Sorani application. The translation of the results of from Kurdish Kurmanji to English is illustrated in Table 18:

Table 18: Translation of the suggestion results from Kurdish Kurmanji to English

| Words in Kurdish Kurmanji | Words in English |
|---|---|
| **em** | this |
| **du** | two |
| **core** | type or kind |
| **babete** | topic |
| **kare** | Job |
| **berhemane** | products |

The provided corpus is vitally important because the algorithm relies on the set of N-grams. The corpus should obtain an adequate sample of word alliances or associations. Furthermore, it has to be small enough to be searched in a fraction of a second. In other words, if the accuracy is poor, the corpus is too small. Table 19 shows the size of the corpus and the performance:

Table 19: Corpus Size Against Accuracy and Performance Results

| Corpus sample size | Accuracy and performance |
| --- | --- |
| Expanding the size of sampled text corpus to 60K | Result: 12.3% accuracy, 38 ms response time |
| Increase the amount of the corpus to 70K | Result: 16.8% accuracy, 40 ms response time, better accuracy. |
| Further increment the corpus to 500K | Result: 17.4% accuracy, 46 ms response time |

From the above results, it can be seen that performance increases with the increment of *N* in the N-gram model. As a result, the straightforward way to assess the used language model is by mapping its accuracy. The result is the ratio of the correctly predicted words. To do so, all the text corpora are utilized to avoid bias in the results. Therefore, the N-grams select only the five most suggested sequences of words detected in the corpus. To measure accuracy, the suggested words are compared with word *n*. In other words, the equation of accuracy is equal to the ratio of correct estimations to total estimations.

$$Accuracy = \frac{correct\ prediction}{total\ prediction} \qquad (22)$$

Table 20 reveals the results of using N-grams, including the accuracy of predictions for the five most likely words.

Table 20: Results for the Accuracy of the Model

| N-grams | Accuracy |
| --- | --- |
| 1-gram | 25.4 |
| 2-gram | 58.6 |
| 3-gram | 72.38 |
| 4-gram | 88.24 |
| 5-gram | 96.3 |

Consequently, a large corpus was utilized to construct the next word prediction system for the Kurdish language. In addition, the N-gram model was used to determine the frequency of the words because the N-gram model is efficient and easily depends on the frequency of the words instead of complex probabilities that affect the performance of the system, as the system proved to be efficient with the N-gram model. In addition, the N-gram model's prediction accuracy is much better than that of other methods because it is easy to use and robust. Moreover, the accuracy of the system shows that utilizing N-grams for this system is adequate to obtain a good result in terms of accuracy. The Stupid BackOff algorithm (SBO) is utilized to predict the likely next words. The algorithm performs better due to the large corpus or dataset and because SBO is ideal for next word prediction. Therefore, SBO prediction has better results than 1 Laplace probability estimation. Consequently, SBO performs better with a large dataset, such as that used for the Kurdish language because SBO estimates the score instead of the probability and therefore does not need to normalize the score or probability. The computation time is costless in SBO, and the accuracy is adequate for a large training dataset. SBO utilizes appropriate frequencies instead of discounting, so it utilizes a backoff score that can be represented as Lambda and equal to 0.4, namely, this approach allows the system to back off to the previous N-gram whenever a prediction is not found. For example, if a prediction cannot be found with the trigram, then it goes back to the bigram and then predicts the next word. This approach is useful because the process will continue and will not stop just because a prediction is not found. As a result, an accurate prediction result is obtained, as shown for the Kurdish language.

## 7. Conclusion

In conclusion, a next word(s) suggestion system has been built for the Sorani and Kuramnji dialects of the Kurdish language. The Kurdish Sorani dialect is different from the English language in terms of morphology, letters, etc. In contrast, the Kurmanji dialect is similar to the English language. The Sorani dialect is written from right to left, unlike the Kurmanji dialect and the English language. The N-gram model was used in the proposed system, and it works well with the Kurdish dialects. In addition, the SBO algorithm was utilized in the system. Whenever the system does not find adequate evidence to predict the next word, it decreases the N-gram, for instance, from the trigram to the bigram. The system predicts the next 5 words to the user. The proposed system adopts several steps: data collection, reading the Kurdish text corpus in R-studio, encoding the text corpus, data cleaning, creating N-grams, saving the N-grams, replacing characters or letters, reading the saved N-grams, and predicting the next words.

The main contributions of this paper are as follows: First, a Kurdish corpus was created, and five grams for the Kurdish language were produced. Researchers can utilize them in their future work. It was more challenging to create this corpus and these grams than it would be for the English language because it is the first time that prediction models for the Kurdish language have been constructed for both the Sorani and Kurmanji dialects. Second, very few Kurdish text corpora with nearly 4,000 words exist. In this work, a new Kurdish text corpus was built with more than 500,000 words for both Kurdish Sorani and Kurmanji. Third, the R-studio software does not show Kurdish Sorani letters properly in objects. When the corpus is encoded to UTF-8, several Kurdish letters are represented in symbols; for example, the (ه) letter in the Kurdish language is represented as (U+0647); this barrier was addressed by replacing the symbols with the corresponding characters inside each N-gram file. After that, the files read into R-studio are listed. Additionally, this system utilizes the N-gram language model, which is suitable for large datasets and useful for calculating frequencies of words instead of complex probabilities. In addition, SBO is used to predict the next words, and it creates good predictions. It is simple to compute with only one parameter, which is Lambda and is equal to 0.4, because it will back off the score when a prediction is not found, for instance, from the quadgram to the trigram. Last, although the Kurdish language is unlike the English language and several issues were faced when building the system, the N-gram language model has an accuracy of 96.3%.

Although a Kurdish text corpus with more than 17 million words exists, there were not adequate resources to use it in the proposed system; a very powerful computer is needed to process a text corpus of that size.

We have no potential conflicts of interest.